\documentclass[10pt,twocolumn,letterpaper]{article}

\def\mL{{\mathcal L}}

\def\0{{\bf 0}}
\def\1{{\bf 1}}

\usepackage{iccv}
\usepackage{times}
\usepackage{epsfig}
\usepackage{graphicx}
\usepackage{amsmath}
\usepackage{amssymb}
\usepackage{booktabs}
\usepackage{multirow}
\usepackage{xcolor}
\usepackage{bbm}
\usepackage{nicefrac}
\usepackage{subfigure}

\usepackage[accsupp]{axessibility}  % Improves PDF readability for those with disabilities.

\usepackage[pagebackref=true,breaklinks=true,letterpaper=true,colorlinks,bookmarks=false]{hyperref}

\iccvfinalcopy % *** Uncomment this line for the final submission

\def\iccvPaperID{2423} % *** Enter the ICCV Paper ID here

\ificcvfinal\fi

\newcommand{\yong}[1]{\textcolor[rgb]{0, 0, 0}{#1}}

\begin{document}

%%%%%%%%% TITLE
\title{Robustifying Token Attention for Vision Transformers}

\author{Yong Guo, David Stutz, Bernt Schiele\\
Max Planck Institute for Informatics, Saarland Informatics Campus\\
{\tt\small guoyongcs@gmail.com, \{david.stutz,schiele\}@mpi-inf.mpg.de}
}

% \and

\maketitle
\ificcvfinal\thispagestyle{empty}\fi

%%%%%%%%% ABSTRACT
\begin{abstract}
       Despite the success of vision transformers (ViTs), 
    they still suffer from significant drops in accuracy in the presence of common corruptions, such as noise or blur. Interestingly, we observe that the attention mechanism of ViTs tends to rely on few important tokens, a phenomenon we call \textit{token overfocusing}. 
    More critically, these tokens are \emph{not} robust to corruptions, often leading to highly diverging attention patterns. In this paper, we intend to alleviate this overfocusing issue and make attention more stable through two general techniques:
    First, our \textbf{Token-aware Average Pooling (TAP)} module encourages the local neighborhood of each token to take part in the attention mechanism.
    Specifically, TAP learns average pooling schemes for each token such that the information of potentially important tokens in the neighborhood can adaptively be taken into account.
    Second, we force the output tokens to aggregate information from a diverse set of input tokens rather than focusing on just a few by using our \textbf{Attention Diversification Loss (ADL)}. We achieve this by penalizing high cosine similarity between the attention vectors of different tokens.
    In experiments, we apply our methods to a wide range of transformer architectures and improve robustness significantly. For example, we improve corruption robustness on ImageNet-C by $2.4\%$ while improving accuracy by $0.4\%$ based on state-of-the-art robust architecture FAN. Also, when fine-tuning on semantic segmentation tasks, we improve robustness on CityScapes-C by $2.4\%$ and ACDC by $3.0\%$. Our code is available at \href{https://github.com/guoyongcs/TAPADL}{https://github.com/guoyongcs/TAPADL}.
\end{abstract}

%%%%%%%%% BODY TEXT
\section{Introduction}

Despite the success of vision transformers (ViTs), their performance still drops significantly on common image corruptions such as 
ImageNet-C \cite{HendrycksICLR2019,wenzel2022assaying,guo2022improving}, adversarial examples \cite{gu2022vision,fu2022patch,shi2021decision,xu2023downscaled}, and out-of-distribution examples as benchmarked in ImageNet-A/R/P~\cite{ZhaoICLR2018,HendrycksICLR2019}. In this paper, we 
examine a key component of ViTs, i.e., the self-attention mechanism, to understand these performance drops. Interestingly, we discover a phenomenon we call \emph{token overfocusing}, where only few important tokens are relied upon by the attention mechanism across all heads and layers. We hypothesize that this overfocusing is particularly fragile to the corruptions on input images and greatly hampers the robustness of the attention mechanism.

\begin{figure}[t]
\begin{center}
   \includegraphics[width=1.0\linewidth]{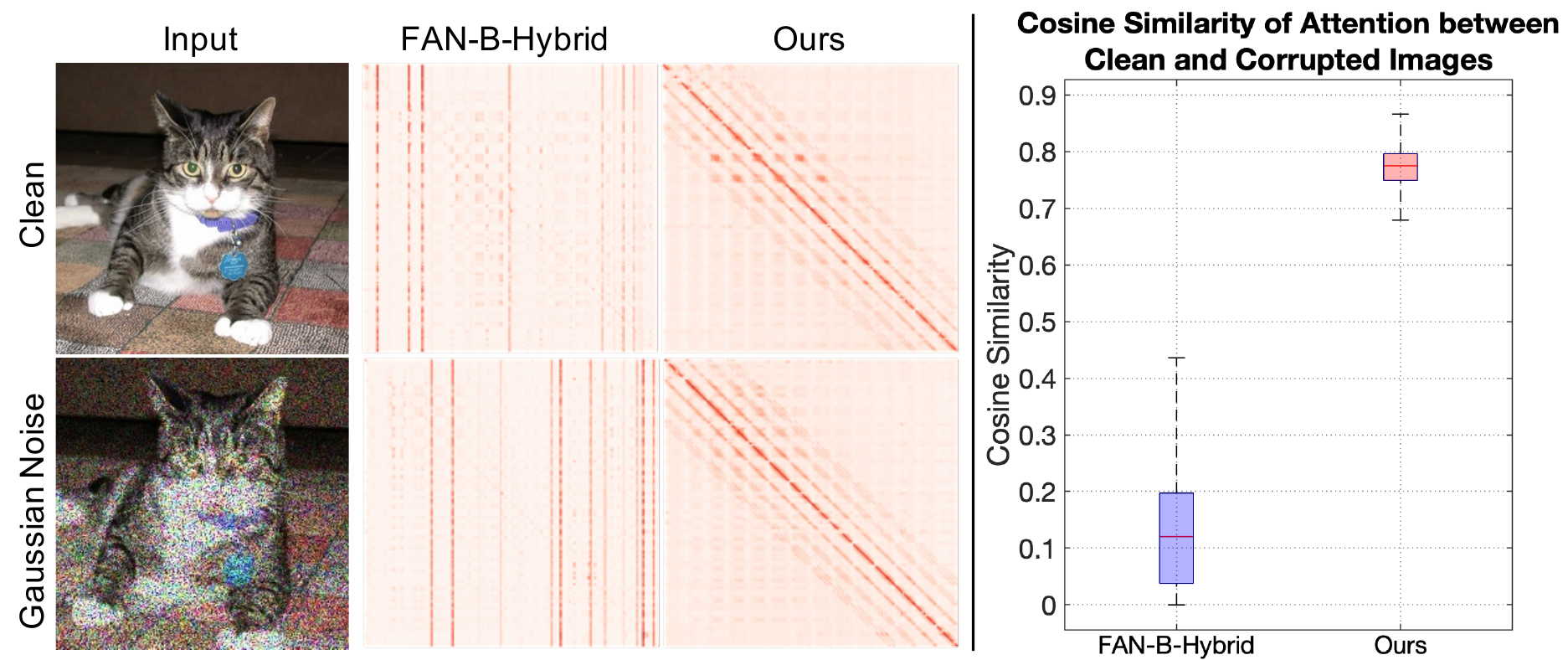}
\end{center}
   \caption{
   Stability against image corruptions in terms of attention visualization (left, a matrix of $196 {\times} 196$) and cosine similarity of attention between clean and corrupted examples (right). 
	\emph{Left}: We average the attention maps across different heads for visualization and show the results of the last layer. We observe that ViTs put too much focus on very few tokens, a phenomenon we call \emph{token overfocusing}. More critically, the attention of the baseline FAN model~\cite{fu2022patch} is fragile to image corruptions, e.g., with Gaussian noise. Our approach, in contrast, alleviates token overfocusing and thereby improves stability of the attention against corruptions.
	\emph{Right}: On ImageNet, we plot the distribution of cosine similarities across all layers (without averaging heads) between clean and corrupted examples. 
	We show that our model yields a significantly higher similarity score than the baseline model.
   }
\label{fig:token_overfocusing}
\end{figure}

Starting from the state-of-the-art robust architecture FAN~\cite{zhou2022understanding}, we exemplarily investigate the last attention layer (see attention of other layers in supplementary). Specifically, Figure~\ref{fig:token_overfocusing} shows a clean and a corrupted input image as well as the corresponding attention maps. These are matrices of $N \times N$, with $N$ being the number of input/output tokens. Here, the $i$-th row indicates which input tokens (columns) the $i$-th output token ``attends'' to -- darker {\color{red}red} indicates higher attention scores. \emph{\textbf{Token overfocusing}} can then, informally, be defined by observing pronounced vertical lines in the attention map: First, each output token (row) focuses on only few important input tokens, ignoring most of the other information. Second, all output tokens seem to focus on the same input tokens, leading to a very low diversity among the attention vectors in different rows. 
We highlight that the overfocusing issue is present throughout the entire ImageNet dataset, and also across diverse architectures (see more examples in Figure~\ref{fig:visual_attention} and supplementary). 
\yong{Moreover, this overfocusing issue has also been observed in existing works. For example, Figure 5 of~\cite{CordonnierICLR2020} observes very few important tokens (in deep red color) in layers 3$\sim$6; Figure 1 of~\cite{fu2022patch} also reports very few important tokens (in yellow color).}
More critically, we find that these important tokens are extremely fragile in the presence of common corruptions. To be specific, when applying Gaussian noise on the input image, the tokens recognized as important change entirely, see Figure \ref{fig:token_overfocusing} (left, second column). 
Quantitatively, this can be captured by computing the cosine similarity between the clean and corrupted attention maps. 
Unsurprisingly, as shown by the \textcolor{blue}{blue box} in Figure~\ref{fig:token_overfocusing} (right), the cosine similarity is indeed extremely low, confirming our initial hypothesis. 
% \yong{
%  We found that this behavior correlates strongly with poor robustness across architectures (including RVT and segmentation models, see Figures II and IV of supplementary) and corruption types (e.g., diverse adversarial conditions in Figure V of supplementary).
% }
This motivates us to robustify the attention by alleviating the token overfocusing issue.

\yong{
To this end, we encourage the attention module to focus on diverse input tokens. This can be achieved by \emph{changing the patterns in both columns and rows} of an attention map, which motivates the two key components of our method.}
% In this paper, we intend to address the token overfocusing issue from two perspectives.
First, 
\yong{when comparing attention columns in Figure~\ref{fig:token_overfocusing}, very few tokens are important, leading to a fragile attention.}
To address this, we encourage output tokens to not only focus on individual input tokens but take into account the local neighborhood around these tokens, in order to make more tokens (columns) contribute meaningful information. 
Intuitively, an individual token itself may not be important but can be enhanced by aggregating the information from potentially important tokens located within its neighborhood.
We achieve this using a learnable average pooling mechanism applied to each input token to aggregate information before computing self-attention.
Second, 
\yong{when comparing attention rows in Figure~\ref{fig:token_overfocusing}, most output tokens focus on the same input tokens. Once these tokens are distorted, the whole attention may fail.}
To address this issue, we seek to diversify the set of input tokens (columns) that the output tokens (rows) rely on. We achieve this using a loss that explicitly penalizes high cosine similarity across rows.
In Figure~\ref{fig:token_overfocusing}, the combination of these techniques leads to more balanced attention across columns and more diverse attention across rows. More critically, these attention maps are more stable in the light of image corruptions. Again, we quantitatively confirm this on ImageNet using the cosine similarity which is  significantly higher.

Overall, we make three key \textbf{contributions}: 
1) we propose a \emph{\textbf{Token-aware Average Pooling (TAP)}} module that encourages the local neighborhood of tokens to participate in the self-attention mechanism. To be specific, we conduct averaging pooling to aggregate information and control it by adjusting the pooling area for each token. 
2) We further develop an \emph{\textbf{Attention Diversification Loss (ADL)}} to improve the diversity of attention received by different tokens, i.e., rows in the attention map. To this end, we compute the attention similarity between rows in each layer and minimize it during training.
3) We highlight that both TAP and ADL can be applied on top of diverse transformer architectures.
In the experiments, the proposed methods consistently improve the model robustness on out-of-distribution benchmarks by a large margin while preserving a competitive improvement of clean accuracy at the same time. 
\yong{
For example, we improve robustness against corruptions and distribution shifts on ImageNet-A/C/P/R by at least $1.5\%$, as shown in Table~\ref{tab:imagenet}.}
In addition, the improvement also generalizes well to other downstream tasks, e.g., semantic segmentation, with an improvement of $2.4\%$ in Table~\ref{tab:segmentation}.

\section{Related Work}
\label{sec:related}

The remarkable performance of ViTs on various learning tasks is largely attributed to the use of self-attention \cite{KhanACM2022}. 
Naturally, the self-attention mechanism has been extended in various aspects, addressing shortcomings such as the heavy reliance on pre-training \cite{touvron2020deit,yuan2021tokens,pan2023improving} or the high computational complexity \cite{liu2021swin,liu2021swinv2,dong2021cswin,ali2021xcit,child2019generating,beltagy2020longformer,wang2020linformer} and adapting the architecture to vision tasks by considering multi-scale transformers \cite{huang2021shuffle,fang2021msg,wang2020axial,wang2021pyramid,wang2021pvtv2,chu2021twins,xu2021coscale,wu2021cvt,huang2021shuffle,wang2021crossformer,chen2021regionvit,yang2021focal}. 
Similar to us, some related works \cite{zhou2021deepvit,CordonnierICLR2020,VigACL2019,naseer2021intriguing,voita2019analyzing} also investigate and visualize the self-attention in order to improve transformer architectures, e.g., to learn deeper transformers \cite{zhou2021deepvit} or prune attention heads \cite{voita2019analyzing}. However, to the best of our knowledge, we are the first to report a token overfocusing issue and link it to poor robustness of ViTs.

There is also a lot of interests in understanding and improving the robustness of ViTs \cite{bhojanapalli2021understanding,bai2021transformers,shi2020robustness,paul2022vision,benz2021robustness,han2022robustify,MahmoodICCV2021}. For example, \cite{mao2021towards} develops a robust vision transformer (RVT) by combining various components to boost robustness, including an improved attention scaling. FAN~\cite{zhou2022understanding} combines token attention and channel attention~\cite{ali2021xcit} and can be considered state-of-the-art. Both approaches rely on modified self-attention mechanisms and can be shown to suffer from token overfocusing. Thus, our techniques can be shown to improve robustness on top of RVT and FAN, respectively. Our TAP approach also shares similarities to recent ideas of trying to introduce locality into the self-attention mechanism \cite{xiao2021early,wu2021cvt,graham2021levit,peng2021conformer,yuan2021incorporating,xiao2021early,wang2021evolving}. For example,  CvT~\cite{wu2021cvt} introduces convolutions into the self-attention architecture to enhance the local information inside tokens. MViTv2~\cite{li2022mvitv2} exploits average pooling to extract local features and improve clean performance. However, all of these strategies use the same aggregation across all tokens, while our approach adaptively chooses an aggregation neighborhood for each token to improve robustness.

\begin{figure*}[t]
\vspace{10 pt}
\begin{center}
   \includegraphics[width=1.0\linewidth]{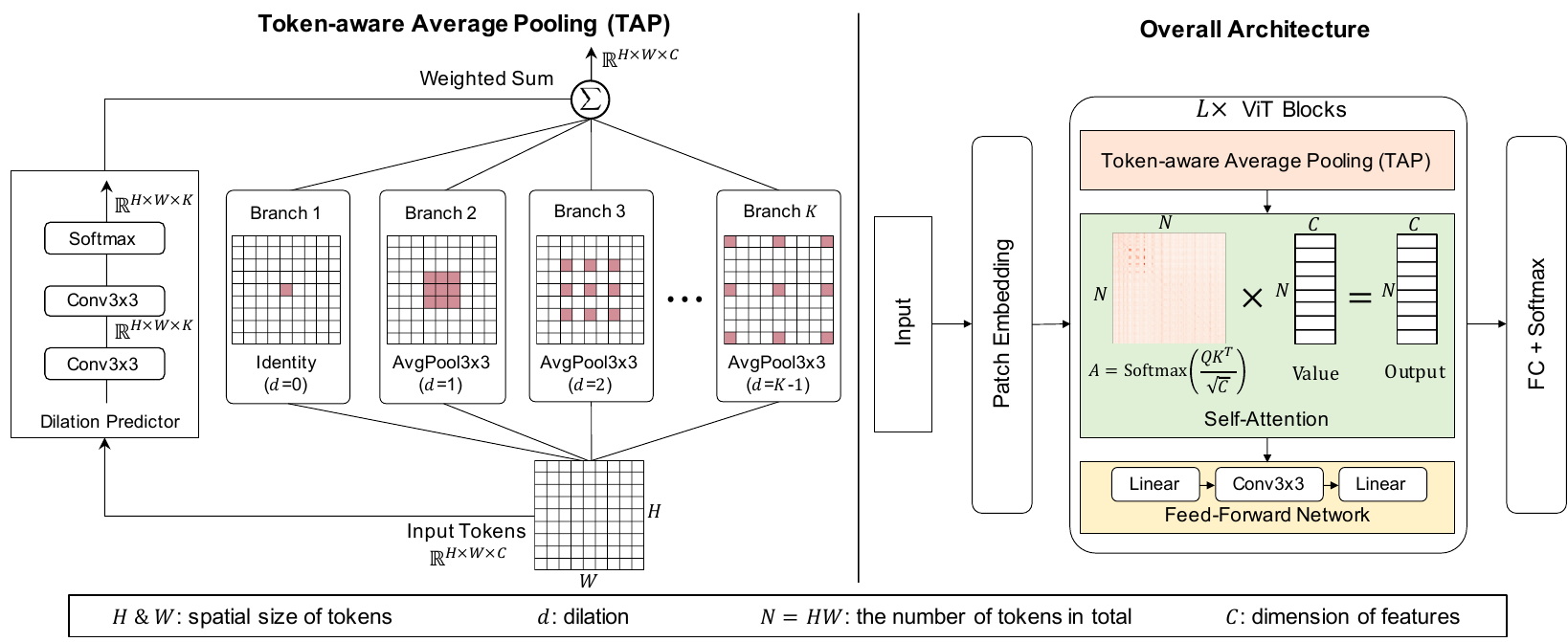}
\end{center}
\vspace{10 pt}
   \caption{The proposed Token-aware Average Pooling (TAP) module (left) and the overall architecture (right). \emph{Left}: In TAP, we introduce $K$ branches to enable tokens to consider different pooling areas and compute the weighted sum over them. Besides, we build a lightweight dilation predictor to learn the weights for different branches. \emph{Right}: We introduce a TAP layer into every basic block to encourage more tokens to be actively involved in the following self-attention mechanism.
   }
\label{fig:tap}
\end{figure*}

\section{Robust Token Self-Attention}

In the following, we focus on the attention mechanism in ViTs, aiming to improve their overall robustness.
To get started, we first describe the observed \emph{token overfocusing issue} where ViTs focus on very few but unstable tokens in detail in Section~\ref{subsec:token_overfocusing}. Then, we propose two general techniques for alleviating this issue: First, in Section~\ref{subsec:tap}, we propose a {\textbf{Token-aware Average Pooling (TAP)}} module that encourages the neighborhood of tokens to participate in the attention mechanism. This is achieved using a learnable pooling area as illustrated in Figure~\ref{fig:tap}. Second, in Section~\ref{subsec:adl}, we develop a new {\textbf{Attention Diversification Loss (ADL)}} to improve the diversity of attention patterns across tokens. Both methods can be applied to most transformer architectures and we will show they greatly improve robustness with negligible training overhead.

\subsection{Token Overfocusing}
\label{subsec:token_overfocusing}

As illustrated in Figure~\ref{fig:token_overfocusing}, we can visualize the self-attention mechanism in each layer as $N \times N$ attention matrix. Here, $N$ is the number of input and output tokens and each entry $(i,j)$ denotes the attention that the $i$-th output token ($i$-th row) puts on the $j$-th input token ($j$-th column) -- deeper \textcolor{red}{red} denoting higher attention scores. To handle multi-head self-attention, we visualize this matrix by averaging across attention heads.

As baseline, Figure~\ref{fig:token_overfocusing} highlights the recent FAN \cite{zhou2022understanding} architecture, showing the attention map of the last layer for example. We observe that the attention is generally very sparse in columns, meaning that most input tokens are not attended to and very few tokens are overfocused. 
More importantly, these ``important'' tokens are often similar across output tokens (rows). We refer to this phenomenon as \emph{token overfocusing}. This leads to problems when facing corruptions such as Gaussian noise: the set of important tokens might change entirely (Figure~\ref{fig:token_overfocusing}, second column). This can be understood as the original tokens not capturing robust information. We can also quantitatively capture this instability by computing  the cosine similarity between clean and corrupted attention maps across all ImageNet images. As shown by the \textcolor{blue}{blue box} in Figure~\ref{fig:token_overfocusing} (right), the baseline model obtains very low cosine similarities, indicating the poor robustness of standard self-attention. 

We found that this phenomenon exists across diverse architectures, e.g., DeiT~\cite{TouvronICML2021} and RVT~\cite{mao2021towards}, and diverse tasks, including both image classification and semantic segmentation (see visual examples in supplementary).
\yong{Moreover, we highlight that our observation is consistent with existing works~\cite{CordonnierICLR2020,pan2022less}. Specifically, these works observe that the middle and deep layers (often with overfocusing issue) tend to capture global information but put focus on very few tokens, e.g., very few deep red tokens in Figure 5 of~\cite{CordonnierICLR2020} and Figure~3~(Stage 4) in~\cite{pan2022less}. This further justifies the existence and importance of the token overfocusing issue.}
To alleviate this issue, we propose two general techniques to robustify self-attention in the remainder of this section.

\subsection{Token-aware Average Pooling (TAP)}
\label{subsec:tap}

In the first part of our method, we seek to encourage more input tokens to participate in the self-attention mechanism, i.e., obtaining more columns with high scores in the attention map. To this end, we encourage each input token to explicitly aggregate useful information from its local neighborhood, in case the token itself does not contain important information. This approach is justified by existing works~\cite{wu2021cvt,peng2021conformer,li2022mvitv2} and the observation that introducing any local aggregation before self-attention consistently improves robustness, see Table~\ref{tab:local_aggregation} (last column).
In fact, these methods apply a fixed convolutional kernel or pooling area to all the tokens. Nevertheless, tokens often differ from each other and each token should require a specific local aggregation strategy. This motivates us to adaptively choose the right neighborhood size and aggregation strategy.

Inspired by this, we enable each token to select an appropriate area/strategy to conduct local aggregation. Specifically, we develop a \textbf{Token-aware Average Pooling (TAP)} that conducts average pooling and adaptively adjusts the pooling area for each token. As shown in Figure~\ref{fig:tap}, we exploit a multi-branch structure that computes the weighted sum over multiple branches, each with a specific pooling area. 
Instead of simply changing the kernel size similar to~\cite{yoo2015multi}, TAP changes the dilation to adjust the pooling area. The main observation behind this is that average pooling with a large kernel, without dilation, leads to extremely large overlaps between adjacent pooling regions and thereby severe redundancy in output tokens. 
This can, for example, be seen in Table~\ref{tab:local_aggregation} where
AvgPool5x5 incurs a large drop in clean accuracy of around 1.2\%.
Similar to \cite{xiao2021early,wu2021cvt,graham2021levit,peng2021conformer}, we also investigated using learnable convolutions, but observed only marginal improvement alongside a significant increase in computational cost.

Based on these observations, we build TAP based on average pooling with diverse dilations:
Without loss of generality, given $K$ branches, we consider the dilation within the range $d {\in} [0,K{-}1]$. Here, $d=0$ implies identity mapping without any computation, i.e., no local aggregation. The maximum dilation determined by $K$ is a hyper-parameter and we find that performance and robustness improvements saturate at around $K = 5$ (see Figure~\ref{fig:k_lambda} top). Within the allowed dilation range, our method includes a lightweight dilation predictor to predict which dilation (i.e., which branch in Figure~\ref{fig:tap}) to utilize. Note that this can also be a weighted combination of multiple $d$. We emphasize that this predictor is very efficient since it reduces the feature dimension (from $C$ to $K$ in Figure~\ref{fig:tap}) such that it adds minimal computational overhead and model parameters. The same approach can also be applied to non-dilated average-pooling where $d$ controls the kernel size, named TAP (multi-kernel). From Table~\ref{tab:local_aggregation}, our TAP greatly outperforms this variant, indicating the effectiveness of using dilations for pooling.

\begin{table}[t]
\begin{center}
\resizebox{1\linewidth}{!}
{
    \begin{tabular}{l|c|c|c|c}
    \toprule
    Model & \multicolumn{1}{c|}{\#FLOPs (G)} & \multicolumn{1}{c|}{\#Params (M)} & \multicolumn{1}{c|}{ImageNet} & \multicolumn{1}{c}{ImageNet-C $\downarrow$} \\
    \hline
    Baseline (FAN-B-Hybrid) &   11.7    &  50.4      &   83.9    & 46.1 \\
    ~~~+~AvgPool3x3 &   11.7    &  50.4      &   83.6    & 45.6 (-0.5) \\
    ~~~+~AvgPool5x5 &   11.7    &  50.4      &   82.7    & 45.5 (-0.6) \\
    ~~~+~Conv3x3 &    17.3   &  79.4     &   84.0    &  45.9 (-0.2) \\
    ~~~+~Conv5x5 &   27.4    &   130.7    &   \textbf{84.4}    & 45.8 (-0.3) \\
    \hline
    ~~~+~TAP (multi-kernel) &   11.8    &   50.7    &   84.1    & 45.5 (-0.6) \\
    ~~~+~TAP (Ours)   &   11.8    &   50.7    &   {84.3}    & \textbf{44.9 (-1.2)} \\
    \bottomrule
    \end{tabular}%
}
\end{center}
  \caption{Comparisons of local aggregation approaches based on FAN-B-Hybrid. We show that conducting average pooling for all the tokens improves the robustness but impedes clean accuracy. Introducing a convolution into each block greatly increases model complexity. In addition, we also compare a variant of our TAP, namely TAP (multi-kernel), that considers multiple kernel sizes for pooling and learns weights for each branch. Our Tap greatly outperforms this variant, indicating the effectiveness of using dilation. Moreover, TAP yields the best tradeoff between accuracy and robustness along with negligible computational overhead. 
  }
  \label{tab:local_aggregation}%
\end{table}%

\begin{table*}[htbp]
\begin{center}
  \resizebox{1.0\textwidth}{!}
  {
    \begin{tabular}{l|c|c|c|cc|cc}
    \toprule
    Method & \#Params (M) & \#FLOPs (G) & {ImageNet $\uparrow$} & {ImageNet-C $\downarrow$} & ImageNet-P $\downarrow$ & {ImageNet-A $\uparrow$} & ImageNet-R $\uparrow$  \\
    \hline
    ConvNeXt-B~\cite{liu2022convnet} & 88.6 & 15.4 & 83.8 & 46.8 & - & 36.7 & 51.3  \\
    % \hline
    % DeiT-B~\cite{TouvronICML2021} & 86.6 & 17.6 & 82.0  & 48.5 & 32.1 & 27.4 & 44.9  \\
    ConViT-B~\cite{d2021convit} & 86.5 & 17.7 & 82.4 & 46.9 & 32.2 & 29.0 & 48.4  \\
    % XCiT-S12~\cite{ali2021xcit} & 26.3 & 4.8 & 81.9 & 51.5 & - & 25.0  & 45.5  \\
    % XCiT-S24~\cite{ali2021xcit} & 47.7 & 9.1 & 82.6 & 49.4 & - & 27.8  & 45.5  \\
    Swin-B~\cite{liu2021swin} & 87.8 & 15.4 & {83.4} & 54.4 & 32.7 & 35.8 & 46.6  \\
    % PVT-Large~\cite{wang2021pyramid} & 61.4 & 9.8 & 81.7 & 59.8 & 39.3 & 26.6 & 42.7  \\
    % PiT-B~\cite{heo2021rethinking} & 73.8 & 12.5 & 82.4 & 48.2 & 36.3 & 33.9 & 43.7   \\
    T2T-ViT\_t-24~\cite{yuan2021tokens} & 64.1 & 15.0 & 82.6 & 48.0 & 31.8 & 28.9 & 47.9  \\
    % RVT-B-RSPC~\cite{guo2023improving} & 91.8 & 17.7 & 82.8 & 45.7 & 31.0 & 32.1 & - \\ 
    RSPC (FAN-B-Hybrid)~\cite{guo2023improving} & 50.5 & 11.7 & 84.2 & 44.5 & 30.0 & 41.1 & - \\
    \hline
    RVT-B~\cite{mao2021towards} & 91.8 & 17.7 & 82.6  & 46.8 & 31.9  & 28.5 & 48.7  \\
    ~~+~TAP & 92.1 & 17.9  & {83.0 (+0.4)}  & {45.5 (-1.3)} & 30.6 (-1.3) & 30.0 (+1.5) & 49.4 (+0.7)  \\
    ~~+~ADL & 91.8 & 17.7 & {82.6 (+0.0)}  & {45.2 (-1.6)} & 30.2 (-1.7) & 30.8 (+2.3) & 49.8 (+1.1)  \\
    ~~+~TAP \& ADL & 92.1 & 17.9 & \textbf{83.1 (+0.5)}  & \textbf{44.7 (-2.1)} & \textbf{29.6 (-2.3)} & \textbf{32.7 (+4.2)} & \textbf{50.2 (+1.5)} \\
    \hline
    FAN-B-Hybrid~\cite{zhou2022understanding} & 50.4 & 11.7 & 83.9  & 46.1 & 31.3  & 39.6 & 52.7 \\
    ~~+~TAP & 50.7 & 11.8 & {84.3 (+0.4)} & {44.9 (-1.2)} & 30.3 (-1.0)  & 41.0 (+1.4) & 53.9 (+1.2)  \\
    ~~+~ADL & 50.4 & 11.7 & {84.0 (+0.1)}  & {44.4 (-1.7)} & 29.8 (-1.5) & 41.4 (+1.8) & 54.2 (+1.5) \\
    ~~+~TAP \& ADL & 50.7 & 11.8 & \textbf{84.3 (+0.4)}  & \textbf{43.7 (-2.4)} & \textbf{29.2 (-2.1)} & \textbf{42.3 (+2.7)} & \textbf{54.6 (+1.9)} \\
    \bottomrule
    \end{tabular}%
    }
\end{center}
  \caption{
  Comparisons on ImageNet and diverse robustness benchmarks. We report the mean corruption error (mCE) on ImageNet-C and mean flip rate (mFR) on ImageNet-P. For these metrics, lower is better. Moreover, we directly report the accuracy on ImageNet-A and ImageNet-R. Based on the considered two baselines, our models consistently improve the accuracy and robustness on diverse benchmarks.
  }
  \label{tab:imagenet}%
\end{table*}%

\subsection{Attention Diversification Loss (ADL)}
\label{subsec:adl}

In the second part of our method, we seek to improve the diversity of attention across output tokens, i.e., encourage different rows in Figure~\ref{fig:token_overfocusing} to attend to different input tokens. 
Based on this objective, we propose an \textbf{Attention Diversification Loss (ADL)} that explicitly reduces the cosine similarity of attention among different output tokens (rows). However, for this approach to work, there are several challenges to overcome. First, computing the cosine similarity between attentions is numerically tricky. For example, if two rows (i.e., output tokens) have very disjoint attention patterns, we expect a low cosine similarity close to 0. However, even for tokens that are not attended to, the attention scores will not be zero. For a large $N$, computing dot product and adding these values up tend to result in a cosine similartiy significantly above zero.
To alleviate this issue, we exploit a thresholding trick to filter out those very small values and only focus on the most important ones. Let $\mathbbm{1}(\cdot)$ be the indication function, and $A_i^{(l)}$ be the attention vector of the $i$-th token (row) in the $l$-th layer. We introduce a threshold $\tau$ (see ablation in Table~\ref{tab:tau}) that depends on the number of tokens $N$, i.e., $\nicefrac{\tau}{N}$. Thus, the attention after thresholding becomes
\begin{equation}\label{eq:attention_threshold}
    \hat A_i^{(l)} = \mathbbm{1} (A_i^{(l)} \geq \nicefrac{\tau}{N} ) \cdot A_i^{(l)}.
\end{equation}

Second, to avoid the quadratic complexity of computing similarities between pairs of $N$ rows,
we approximate it by computing the cosine similarity between each individual attention vector $\hat A_i^{(l)}$ with the average attention $\bar A^{(l)} := \frac{1}{N} \sum_{i=1}^N \hat A_i^{(l)}$. When considering a model with $L$ layers, we average the ADL loss across all the layers by: 
\begin{equation}
    \mL_{\rm ADL} {=} \frac{1}{L} \sum_{l=1}^{L} \mL_{\rm ADL}^{(l)},~~~
    \mL_{\rm ADL}^{(l)} {=} \frac{1}{N} \sum_{i=1}^{N} \frac{\hat A_i^{(l)} \bar A^{(l)}} {\|\hat A_i^{(l)}\| \| \bar A^{(l)} \|}.
\end{equation}
In practice, we combine our ADL with the standard cross-entropy (CE) loss and introduce a hyper-parameter $\lambda$ (see ablation in Figure~\ref{fig:k_lambda}) to control the importance of ADL:
\begin{equation}
\label{eq:objective}
    \mL = \mL_{\rm CE} + \lambda   \mL_{\rm ADL}.
\end{equation}
We highlight that our ADL can be applied to boost the robustness on diverse tasks, including image classification and semantic segmentation (see Table~\ref{tab:imagenet} and Table~\ref{tab:segmentation}).

\begin{table*}[htbp]
\begin{center}
  \resizebox{1\textwidth}{!}
  {
    \begin{tabular}{l|c|ccc|cccc|cccc|cccc}
    \toprule
    \multicolumn{1}{l|}{\multirow{2}[0]{*}{Method}} &    \multirow{2}[0]{*}{mCE}   & \multicolumn{3}{c|}{Noise} & \multicolumn{4}{c|}{Blur}      & \multicolumn{4}{c|}{Weather}   & \multicolumn{4}{c}{Digital} \\
          &  & \multicolumn{1}{l}{Gaussian} & \multicolumn{1}{l}{Shot} & \multicolumn{1}{l|}{Impulse} & \multicolumn{1}{l}{Defocus} & \multicolumn{1}{l}{Glass} & \multicolumn{1}{l}{Motion} & \multicolumn{1}{l|}{Zoom} & \multicolumn{1}{l}{Snow} & \multicolumn{1}{l}{Frost} & \multicolumn{1}{l}{Fog} & \multicolumn{1}{l|}{Brightness} & \multicolumn{1}{l}{Contrast} & \multicolumn{1}{l}{Elastic} & \multicolumn{1}{l}{Pixelate} & \multicolumn{1}{l}{JPEG} \\
    \hline
    FAN-B-Hybrid & 46.2  & 40.12 & 39.27 & 36.80  & 51.58 & 63.96 & 47.53 & 54.98 & 40.24 & 43.96 & 36.98 & 36.68 & 34.17 & 61.59 & 53.25 & 51.86 \\
    ~~+~TAP   & 44.9 & 36.02 & 36.26 & 34.16 & 52.72 & 65.07 & 45.73 & 54.90  & 39.75 & 42.39 & 35.68 & 37.38 & 33.21 & 62.75 & 48.85 & 49.46 \\
    ~~+~ADL    & 44.3 & 35.61 & 35.55 & 33.51 & \textbf{50.80}  & 64.27 & 45.47 & 54.47 & \textbf{38.06} & 40.46 & 37.92 & 36.99 & {32.70}  & \textbf{61.45} & 47.78 & \textbf{49.00} \\
    ~~+~TAP \& ADL & \textbf{43.7} & \textbf{33.87} & \textbf{34.24} & \textbf{32.04} & 51.29 & \textbf{61.51} & \textbf{44.76} & \textbf{54.39} & 38.14 & \textbf{40.12} & \textbf{35.27} & \textbf{36.43} & \textbf{32.25} & 62.12 & \textbf{46.78} & 49.55 \\
    \bottomrule
    \end{tabular}%
    }
\end{center}
  \caption{Comparisons of corruption error (lower is better) on individual corruption type of ImageNet-C based on FAN-B-Hybrid. Combining TAP and ADL together yields the best results on most of the corruption types. 
  }
  \label{tab:imagenetc_individual}%
\end{table*}%

\section{Experiments}

We conduct extensive experiments to verify our method on both image classification and semantic segmentation tasks. In Section~\ref{subsec:classification}, we first train classification models on ImageNet~\cite{deng2009imagenet} and demonstrate that our models obtain significant improvement on various robustness benchmarks, including ImageNet-A~\cite{ZhaoICLR2018}, ImageNet-C~\cite{HendrycksICLR2019}, ImageNet-R~\cite{HendrycksARXIV2020}, and ImageNet-P~\cite{HendrycksICLR2019}. 
Then, in Section~\ref{subsec:segmentation}, we take our best pre-trained model and further finetune it on Cityscapes~\cite{cordts2016cityscapes} for semantic segmentation. In practice, our models greatly improve mIoU on two popular robustness benchmarks, including Cityscapes-C~\cite{michaelis2019benchmarking} and ACDC~\cite{sakaridis2021acdc}, along with competitive performance on clean data. 
Both the code and pretrained models will be available soon.

\subsection{Results on Image Classification}
\label{subsec:classification}

In this experiment, we build our method on top of two state-of-the-art robust architectures: RVT~\cite{mao2021towards} and FAN~\cite{zhou2022understanding} with the ``Base'' model size, i.e., RVT-B and FAN-B-Hybrid. We train the models on ImageNet and evaluate them on several robustness 
% \david{robustness benchmarks?} 
benchmarks. We closely follow the settings of RVT and FAN for training. Specifically, we train the models using the same augmentation schemes and adopt the batch size of 2048. 
We set the learning rate to $2 {\times} 10^{-3}$ and train all the models for 300 epochs. In all the experiments, by default, we set $K=4$ and $\lambda=1$ to train our models.
To evaluate robustness, we consider several robustness benchmarks, including ImageNet-A~\cite{ZhaoICLR2018}, ImageNet-C ~\cite{HendrycksICLR2019}, ImageNet-R~\cite{HendrycksARXIV2020}, and ImageNet-P~\cite{HendrycksICLR2019}.
Note that, we report the mean corruption error (mCE) on ImageNet-C and mean flip rate (mFR) on ImageNet-P. For both metrics, lower is better.
Empirically, we demonstrate that using either TAP or ADL individually is able to improve the robustness. When combining them together, our models outperform the baselines by a larger margin and the performance improvement generalizes well to diverse architectures (see Table~\ref{tab:diverse_architecture}).

\subsubsection{Comparisons on ImageNet}

As shown in Table~\ref{tab:imagenet}, 
compared with the strong baselines RVT and FAN, our models consistently improve the robustness on ImageNet-C by ${>}2.1\%$ and also yield comparable improvement on other robustness benchmarks, including ImageNet-A/R/SK. Moreover, our models also obtain a competitive improvement in terms of clean accuracy on ImageNet. For example, we improve the accuracy by ${>}0.4\%$ on both considered baseline architectures. More critically, we highlight that these improvements only come with negligible computational cost in terms of both the number of parameters and the number of floating-point operations (FLOPs).
In addition, we also report the detailed corruption error on individual corruption types of ImageNet-C based on FAN-B-Hybrid. From Table~\ref{tab:imagenetc_individual}, our best model (combining TAP and ADL together) obtains the best results on most of the corruption types. It is worth noting that our model is particularly effective against noise corruptions, e.g., yielding a large improvement of $6.25\%$ on Gaussian noise corruption. Overall, these experiments indicate that robustifying attention consistently improves robustness across different architectures and benchmarks.

\subsubsection{Attention Stability and Visualization Results}
\label{subsubsec:attention_stability}

We demonstrate that our models greatly improve attention stability against corruptions both qualitatively and quantitatively. Interestingly, in each layer, our models obtain a higher attention diversity among different attention heads.

\begin{figure}[t]
\begin{center}
   \includegraphics[width=1.0\linewidth]{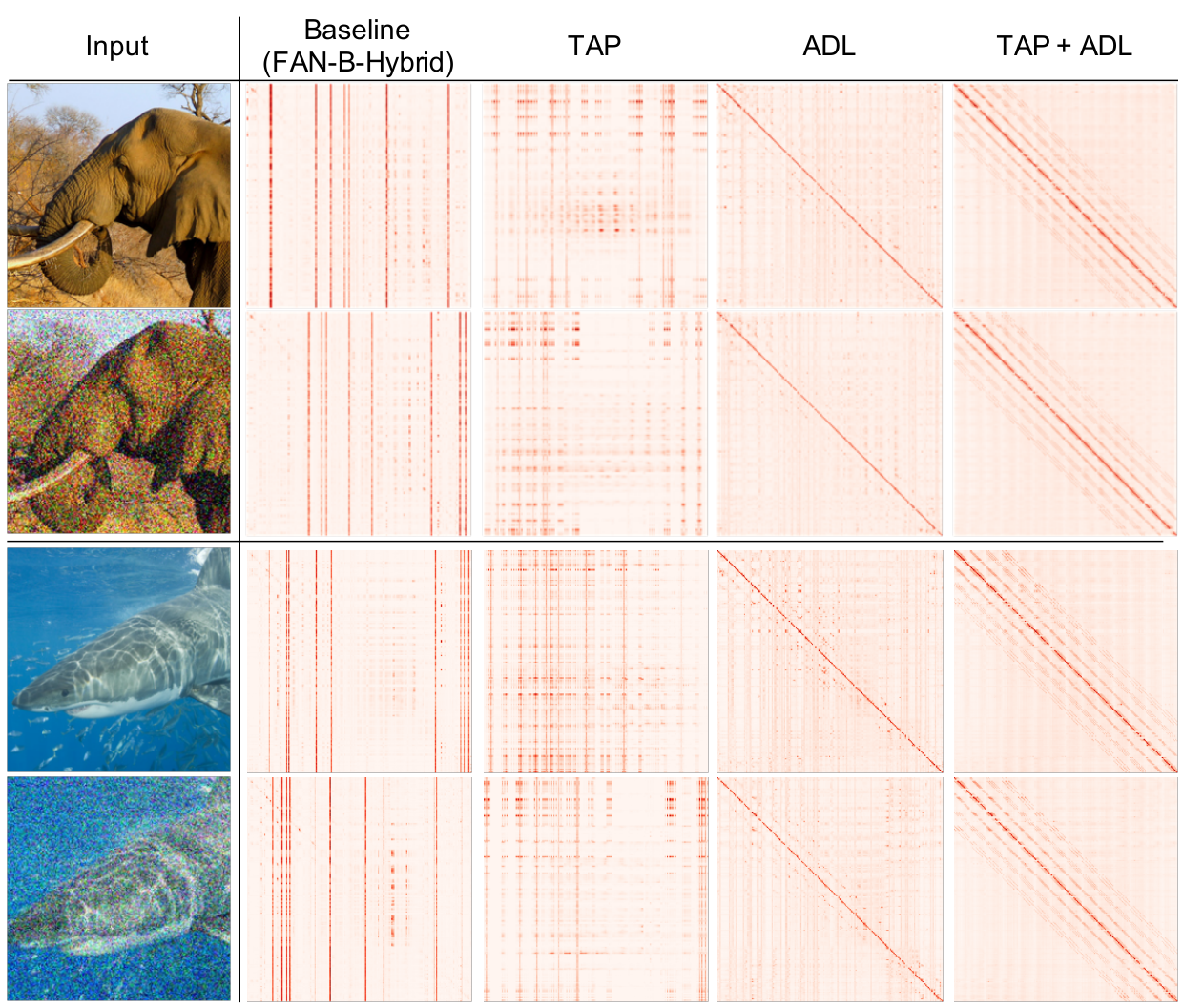}
\end{center}
   \caption{Comparisons of attention maps among different models. Compared with the baseline model, our TAP alleviates the overfocusing issue by encouraging the tokens surrounding the most important ones to have higher attention scores. When only using ADL, we obtain an attention map that follows the diagonal pattern, i.e., preserving the token itself while aggregating information from other tokens. Apparently, the attention rows are different from each other, fulfilling our goal of reducing similarity between rows. When combining TAP and ADL together, the diagonal pattern is further expanded to the nearby areas thanks to the TAP module.}
\label{fig:visual_attention}
\end{figure}

\noindent
\textbf{Attention stability.}
In Figure~\ref{fig:visual_attention} we first visualize how much the attention would be changed when facing image corruptions, e.g., Gaussian noise.
We take FAN-B-Hybrid as the baseline and compare our best model with two variants that only contain TAP and ADL individually. Across diverse examples, the baseline model incurs a severe token overfocusing issue that it puts too much focus on very few tokens and comes with a significant attention shift when facing corruptions. With the help of our local aggregation module TAP, our TAP model assigns the attention to more tokens surrounding some important ones, alleviating the token overfocusing issue to some extent. Nevertheless, we still observe an attention shift between clean and corrupted examples. When training models with our ADL, the attention follows a diagonal pattern such that tokens aggregate information from the others while retaining most of the information from itself. We highlight that this diagonal pattern is somehow similar to the residual learning~\cite{HeCVPR2016} that additionally learns a residual branch while keeping the features unchanged using an identity shortcut. Clearly, the attention becomes much more stable against corruptions. When combining TAP and ADL together, we further encourage the diagonal pattern to expand within a local region. In this way, each token puts more focus on its neighborhood beyond itself and thus obtains stronger features. Quantitatively, we also evaluate the attention stability by computing cosine similarity of attention between clean and corrupted examples across the whole ImageNet. From Figure~\ref{fig:attn_stability}, TAP yields higher attention stability than the baseline model. Thanks to the diagonal pattern in attention, the model with ADL greatly improves the similarity score by a large margin, indicating that the attention is very stable against corruptions. When combining both TAP and ADL, we can further improve attention stability.

\begin{figure*}[t]
\begin{center}
   \includegraphics[width=1.0\linewidth]{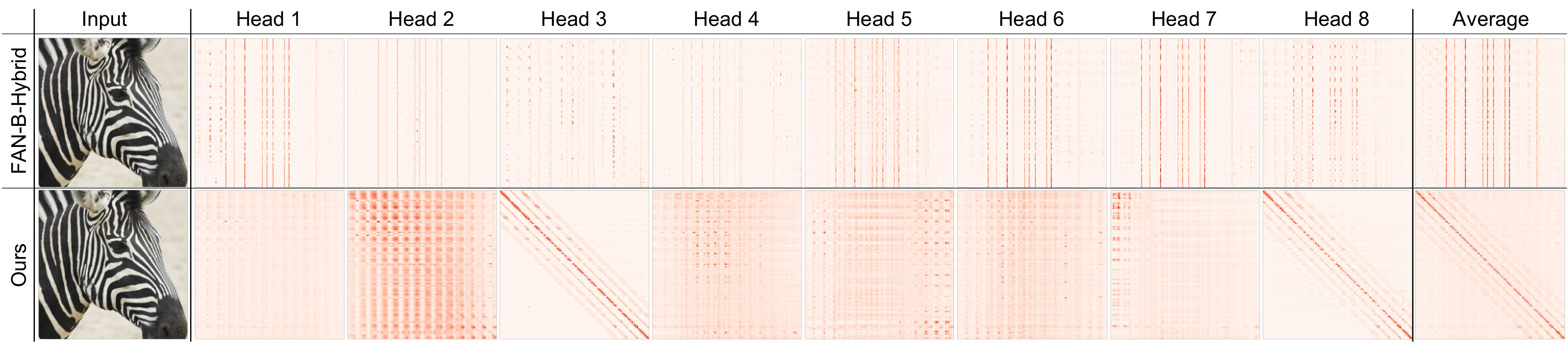}
\end{center}
   \caption{
   Attention maps of different attention heads in the last layer. For the baseline model, the most important tokens are often shared across different heads. In our model, two heads have the attention with the diagonal pattern and the other heads have specific patterns to extract different features, yielding higher attention diversity among heads (see quantitative results in Section~\ref{subsubsec:attention_stability}).}
\label{fig:attention_head}
\end{figure*}

\begin{figure}[t]
\begin{center}
   \includegraphics[width=1.0\linewidth]{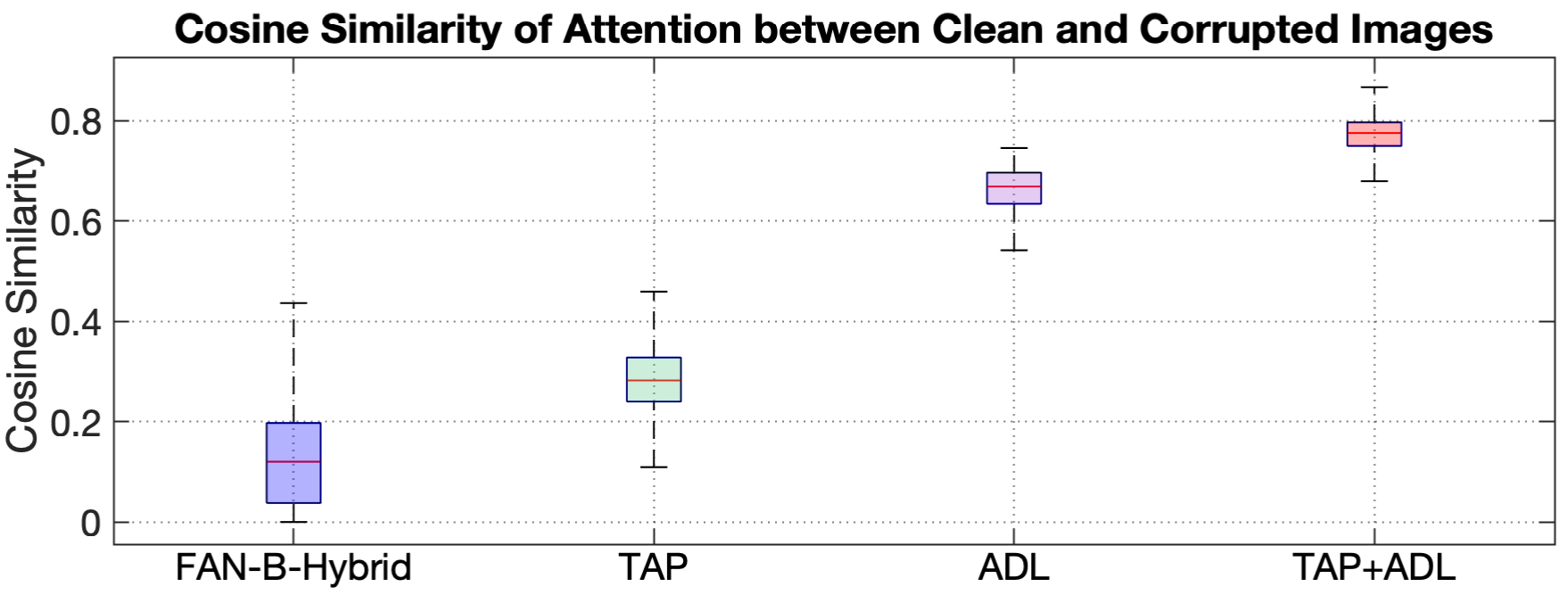}
\end{center}
   \caption{Distributions of  cosine similarity of intermediate attention maps between clean and corrupted examples (e.g., with Gaussian noise) on ImageNet. We demonstrate that either TAP or ADL is able to improve the stability of attention independently. When combining them together, we further improve the stability/similarity of attention against corruptions.}
\label{fig:attn_stability}
\end{figure}

\begin{figure*}[t]
\begin{center}
   \includegraphics[width=1.0\linewidth]{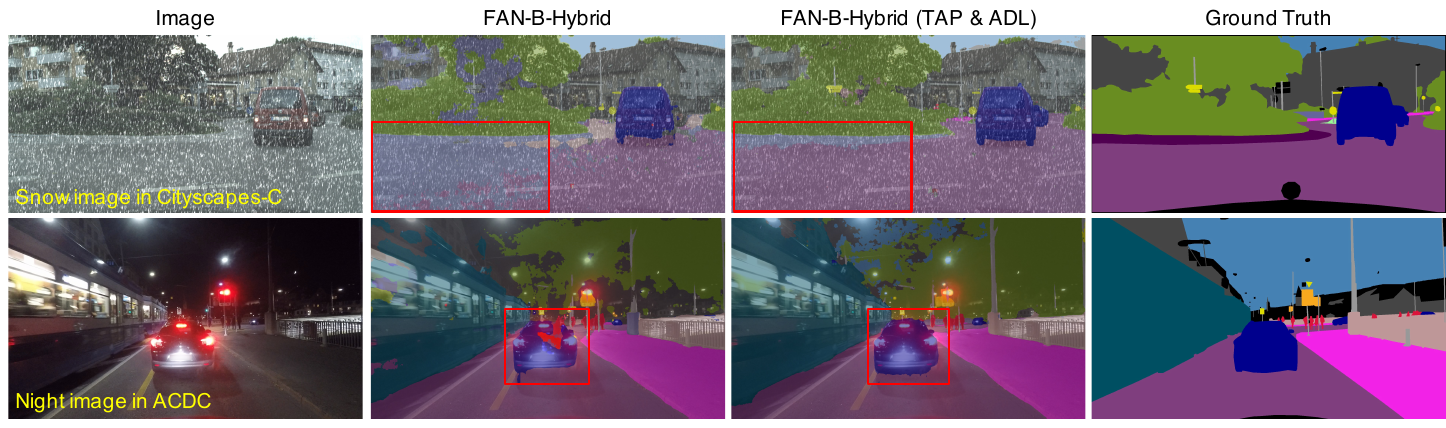}
\end{center}
\vspace{-7 pt}
   \caption{Visual comparisons of segmentation results. When facing image corruptions or adverse conditions, the baseline FAN-B-Hybrid model fails to detect some important objects (e.g., road in the first example) or mistakenly recognizes a part of car as a rider (in the second example). By contrast, our model is much more robust against these corruptions and adverse conditions.}
\label{fig:visual_segmentation}
\end{figure*}

\noindent
\textbf{Attention of each head and attention diversity.} In this part, we further investigate the attention map in each individual head. As shown in Figure~\ref{fig:attention_head}, for the baseline model, the most important tokens are always the most important ones in almost all the heads, resulting in a very low diversity of attention among different heads. By contrast, in our model, only two heads follow the diagonal pattern and the other six heads have different attention patterns from each other. We highlight that our attention is a combination of both local and global filters w.r.t. different heads. Specifically, these two diagonal heads can be regarded as local filters to extract local information since the diagonal pattern encourages tokens to aggregate information within their local neighborhood. As for the other six heads, the attention is distributed across the whole map and thus they can be regarded as global filters. From the visualization result in Figure~\ref{fig:attention_head}, our model has a higher diversity of attention among different heads. To quantify this, following the previously discussed approach, we directly compute the cosine similarity of attention maps between any two heads in each layer (see detailed computation method in supplementary). In other words, the lower the similarity score is, the higher the attention diversity will be. In practice, the baseline model obtains a high similarity score of 0.63 when averaging across the whole ImageNet, indicating a very low attention diversity among different heads. By contrast, our model yields a lower similarity of 0.27, which is consistent with the previous observation that our model produces diverse attention across different heads.

\subsection{Results on Semantic Segmentation}
\label{subsec:segmentation}

In this part, we further apply our method to semantic segmentation tasks. We train the models on Cityscapes~\cite{cordts2016cityscapes} and evaluate the robustness on two popular benchmarks, including Cityscapes-C~\cite{michaelis2019benchmarking} and ACDC~\cite{sakaridis2021acdc}. Specifically, Cityscapes-C contains 16 corruption types which can be
divided into 4 categories: noise, blur, weather, and digital. ACDC collects the images with adverse conditions,
including night, fog, rain, and snow. In this paper, we report mIoU across diverse datasets. During training, we follow the same settings of SegFormer~\cite{xie2021segformer} to train our models.
We demonstrate that our TAP and ADL also generalize well to segmentation tasks and significantly improve robustness.

\subsubsection{Quantitative Comparisons}

As shown in Table~\ref{tab:segmentation},
compared with the considered baseline model, using either our TAP or ADL individually can greatly improve the robustness on Cityscapes-C and ACDC. With the help of our effective local aggregation module TAP, we highlight that we also obtain a promising improvement of 0.5\% mIoU on clean data. When combining TAP and ADL together, we further improve the performance and obtain a larger improvement of 2.4\% and 3.0\% on Cityscapes-C and ACDC, respectively. Moreover, our best model also significantly outperforms several popular segmentation models with comparable model size. Overall, these results indicate that the proposed two techniques not only work for image classification but also generalize well to semantic segmentation tasks.

\begin{table}[t]
\begin{center}
\resizebox{1\linewidth}{!}
{
    \begin{tabular}{l|c|cc}
    \toprule
    Model  & Cityscapes $\uparrow$ & Cityscapes-C $\uparrow$ & ACDC $\uparrow$ \\
    \hline
    % DeepLabv3+ (R50)~\cite{chen2017rethinking} & 25.4 & 76.6 & 36.8 \\
    DeepLabv3+ (R101)~\cite{chen2017rethinking}  & 77.1 & 39.4 & 41.6 \\
    % DeepLabv3+ (X65)~\cite{chen2017rethinking} & 22.8 & 78.4 & 42.7 \\
    % DeepLabv3+ (X71)~\cite{chen2017rethinking} &   -    & 78.6 & 42.5 \\
    % \hline
    ICNet~\cite{zhao2018icnet} & 65.9 & 28.0 & - \\
    % FCN8s~\cite{long2015fully} & 50.1 & 66.7 & 27.4 \\
    DilatedNet~\cite{yu2015multi}   & 68.6 & 30.3 & - \\
    % ResNet38 & 18.0   & 77.5 & 32.6 \\
    % PSPNet~\cite{zhao2017pyramid} & 13.7 & 78.8 & 34.5 \\
    % ConvNeXt-T~\cite{liu2022convnet} & 29.0 & 79.0 & 54.4 \\
    % \hline
    
    Swin-T~\cite{liu2021swin}  & 78.1 & 47.3 & 56.3 \\
    SETR~\cite{zheng2021rethinking}  & 79.5 & 63.1 & 60.2 \\
    % Segformer-B0~\cite{xie2021segformer} & 3.4  & 76.2 & 48.8 \\
    % Segformer-B1~\cite{xie2021segformer} & 13.1 & 78.4 & 52.7 \\
    % Segformer-B2~\cite{xie2021segformer} & 24.2 & 81.0 & 59.6 & 56.2 \\
    Segformer-B5~\cite{xie2021segformer}  & 82.4 & 65.8 & 62.0 \\
    \hline
    FAN-B-Hybrid~\cite{zhou2022understanding}  & 82.2 & 67.3 & 60.6 \\
    ~~+~TAP  & 82.7 (+0.5) & 69.2 (+1.9) & 62.7 (+2.1) \\
    ~~+~ADL  & 82.4 (+0.2) & 69.4 (+2.1) & 63.1 (+2.5) \\
    ~~+~TAP \& ADL  & \textbf{82.9 (+0.7)} & \textbf{69.7 (+2.4)} & \textbf{63.6 (+3.0)}\\
    \bottomrule
    \end{tabular}%
}
\end{center}
  \caption{
  Comparisons of semantic segmentation models on Cityscapes validation set, Cityscapes-C, and ACDC test set. Both our TAP and ADL greatly improve the robustness. We can further improve the results when combining TAP and ADL together.}
  \label{tab:segmentation}%
\end{table}%

\subsubsection{Visual Comparisons}

In this part, we compare the visualization results of the predicted segmentation masks based on examples of diverse robustness benchmarks. As shown in Figure~\ref{fig:visual_segmentation}, for the first example with snow corruptions, the baseline model cannot detect a large region of road (highlighted by the red box), which poses potential risks when applied in some real-world applications, e.g., autonomous driving. Moreover, in the second example with night conditions, the baseline model recognizes a part of the car as a rider and introduces a lot of artifacts in the predicted mask.
By contrast, our model is much more robust and is able to accurately detect most parts of the road and the car in both cases. We highlight that our superiority of robustness can be observed on most examples of the considered benchmarks. Please refer to more visual comparisons in supplementary.

\section{Analysis and Discussions}

In the following, we present further ablation experiments and discussions.
In Section~\ref{subsec:diverse_architecture}, we demonstrate that the proposed two methods are general techniques and can be applied on top of diverse transformer architectures.
In Section~\ref{subsec:hyperparameters}, we study the impact of the number of branches $K$ in TAP and the weight of ADL in the training loss.
\yong{
In Section~\ref{subsec:allocate_tap}, we investigate the strategy of how to allocate TAP layers. In practice, uniformly allocating a TAP into every block yields the best results. In Section~\ref{subsec:threshold}, we study the effect of the attention threshold $\tau$ used in Eqn.~(\ref{eq:attention_threshold}).
}

\subsection{Effectiveness on Diverse Architectures}
\label{subsec:diverse_architecture}

Besides RVT and FAN, we additionally apply our methods on top of more transformer architectures, including DeiT~\cite{TouvronICML2021} and Swin~\cite{liu2021swin}. 
In this experiment, we report the accuracy on ImageNet and robustness in terms of mCE (the lower the better) on ImageNet-C.
As shown in Table~\ref{tab:diverse_architecture}, based on DeiT-B, we greatly improve the corruption robustness by reducing the mCE by 1.9\% and yield a promising improvement of 0.4\% on clean data. As for Swin-B, we obtain a similar observation that our methods are particularly effective in improving corruption robustness, reducing mCE from 54.4\% to 51.9\%. These results indicate that our methods can generalize well across diverse architectures.

\subsection{Impact of Hyperparameters $K$ and $\lambda$}
\label{subsec:hyperparameters}

We conduct ablations on ImageNet-C to study the impact of two hyperparameters of our methods, including the number of branches $K$ in TAP and the weight of ADL.

\begin{figure}[t]
\begin{center}
\includegraphics[width=0.77\linewidth]{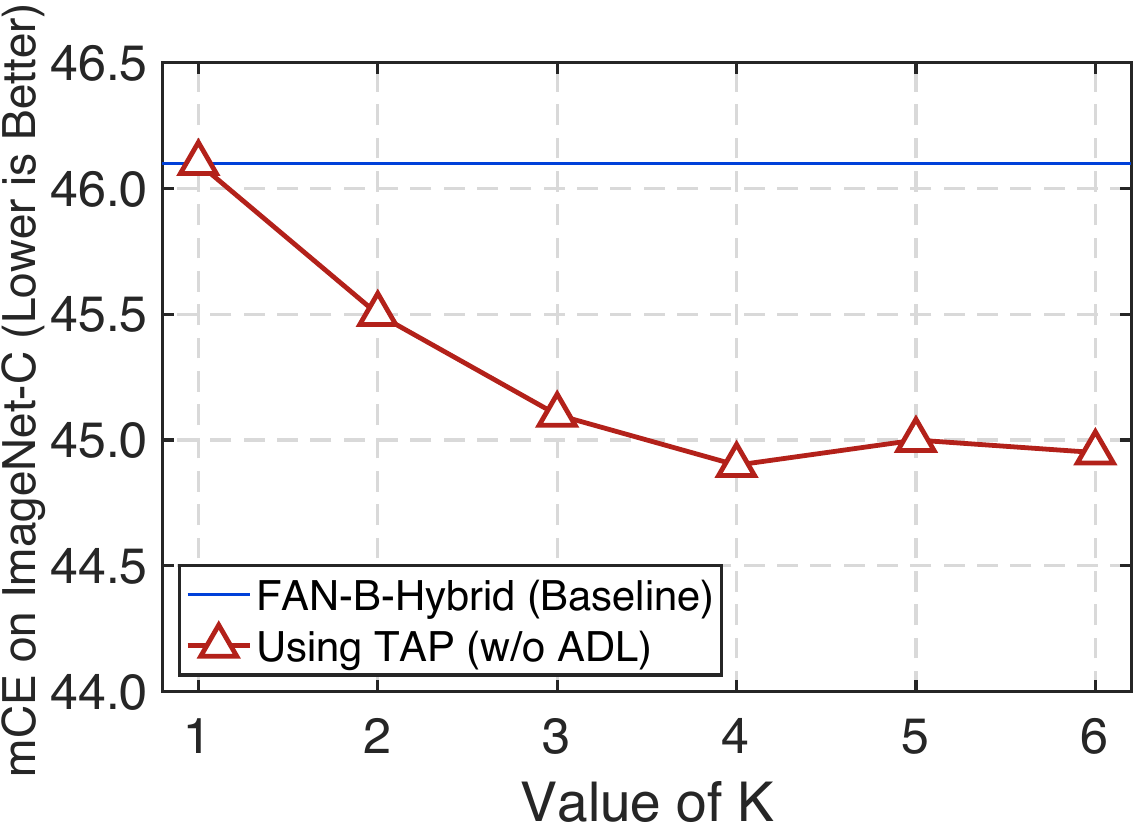} \\
~~\includegraphics[width=0.79\linewidth]{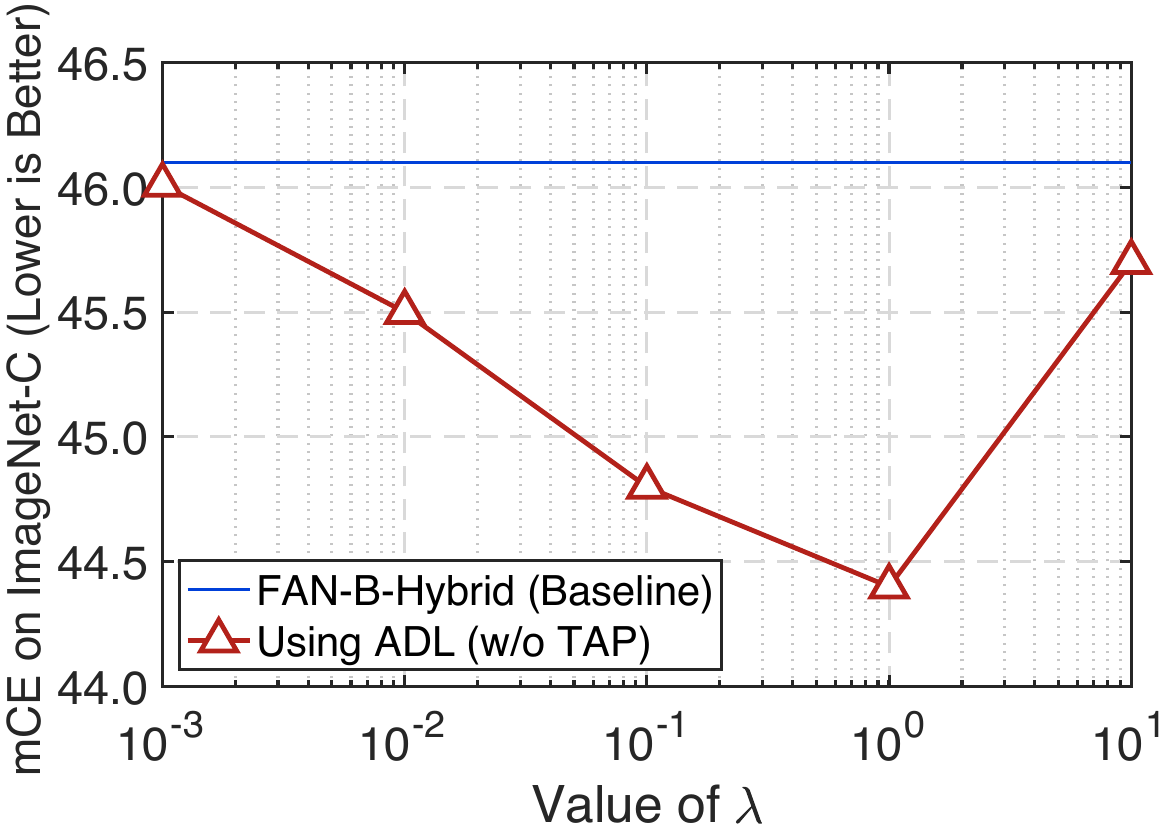}
\end{center}
   \caption{Robustness in terms of mean corruption error (mCE, lower is better) on ImageNet-C against the number of branches $K$ (top) and the importance of our ADL loss $\lambda$ (bottom). \emph{Top:} When only introducing TAP without ADL, our model consistently outperforms the baseline model when increasing the value of $K$ and yields the best result with $K=4$. \emph{Bottom:} When using ADL to train the model (without TAP), we observe that a too small or too large $\lambda$ reduces the benefit of our method. In practice, $\lambda=1$ performs best in most cases.}
\label{fig:k_lambda}
\end{figure}

\noindent
\textbf{The number of branches $K$}.
As detailed in Figure~\ref{fig:tap}, we build our TAP with $K$ branches to enable tokens to consider diverse pooing areas. In fact, the value of $K$ is an important factor for the performance of our method. It is worth noting that $K=1$ is essentially equivalent to the baseline model without TAP. As shown in Figure~\ref{fig:k_lambda} (top), we greatly improve the robustness with a lower mCE score on ImageNet-C when gradually increasing $K$ from 1 to 4. If we further increase $K$, learning weights for too many candidate pooling areas (dilations) for each token becomes increasingly difficult and we cannot observe a significant improvement. 
Although additional branches only introduce minimal overhead in terms of model size and computational complexity, a large $K$ would inevitably require a larger memory footprint. Thus, we choose $K=4$ to obtain the best results at the minimal cost of extra memory footprint.

\begin{table}[t]
\vspace{5 pt}
\begin{center}
\resizebox{0.93\columnwidth}{!}
  {
    \begin{tabular}{l|c|c}
    \toprule
    Method & {ImageNet} & {ImageNet-C (mCE) $\downarrow$}  \\
    \hline
    DeiT-B~\cite{TouvronICML2021} &  82.0  & 48.5  \\
    ~~+~TAP \& ADL & \textbf{82.4 (+0.4)}  & \textbf{46.6 (-1.9)}  \\
    \hline
    Swin-B~\cite{liu2021swin} &  83.4  & 54.4  \\
    ~~+~TAP \& ADL & \textbf{84.0 (+0.6)}  & \textbf{51.9 (-2.5)}  \\
    \hline
    RVT-B~\cite{mao2021towards} &  82.6  & 46.8  \\
    ~~+~TAP \& ADL & \textbf{83.1 (+0.5)}  & \textbf{44.7 (-2.1)}  \\
    \hline
    FAN-B-Hybrid~\cite{zhou2022understanding} & 83.9  & 46.1 \\
    ~~+~TAP \& ADL  & \textbf{84.3 (+0.4)}  & \textbf{43.7 (-2.4)} \\
    \bottomrule
    \end{tabular}%
    }
\end{center}
\vspace{3 pt}
  \caption{Results on top of diverse architectures. We report accuracy and mean corruption error (mCE) on ImageNet and ImageNet-C, respectively. Our method consistently improves robustness and accuracy across different architectures.}
  \label{tab:diverse_architecture}%
\end{table}%

\noindent 
\textbf{Weight of ADL $\lambda$}.
In Figure~\ref{fig:k_lambda} (bottom), we change the value of $\lambda$ in Eqn.~(\ref{eq:objective}).
In practice, a larger $\lambda$ encourages models to diversify the attention among different rows in the attention map more aggressively.
Given a set of values $\lambda \in \{0.001, 0.01, 0.1, 1, 10\}$, we gradually improve the robustness (reduce mCE score) until $\lambda = 1$. When considering an even larger $\lambda=10$, we observe a significant performance drop since a too large $\lambda$ for ADL may hamper the standard cross-entropy loss during training. In this paper, we set $\lambda=1$ and it generalizes well across all the considered architectures and learning tasks.

\subsection{Strategies of Allocating TAP Layers} \label{subsec:allocate_tap}

\yong{
In this part, we explicitly investigate how to allocate TAP layers into a transformer model.
As discussed in Section~\ref{subsec:tap}, TAP is a learnable module that can adjust itself to fit different layers. Specifically, when overfocusing is not severe, TAP can be reduced to identity mapping by setting the weight of the first branch to 1 in Figure~\ref{fig:tap}. 
Empirically, in Table~\ref{tab:nonuniform_tap} (without ADL), when introducing TAP into shallow layers (first 30\% layers) where overfocusing is not severe, we observe marginal improvement on ImageNet-C. However, for middle and deep layers (the rest 70\% layers with overfocusing), we obtain large improvement, similar to uniformly allocating TAP. To avoid manually selecting layers, we suggest allocating TAP into every block.
}

\begin{table}[t]
\begin{center}
\resizebox{1\columnwidth}{!}
    {
    \begin{tabular}{c|c|ccc}
    \hline
    Method & Baseline & \multicolumn{1}{l}{TAP (Shallow)} & \multicolumn{1}{l}{TAP (Middle+Deep)} & \multicolumn{1}{l}{TAP (All)}\\
    \hline
    ImageNet $\uparrow$ &   83.9    &  84.1 (+0.2)     & \textbf{84.3 (+0.4)} & \textbf{84.3 (+0.4)} \\
    ImageNet-C $\downarrow$ &   46.1    &   45.7 (-0.4)    & 45.0 (-1.1)  & \textbf{44.9 (-1.2)} \\
    \hline
    \end{tabular}%
    }
\end{center}
  \caption{Comparisons of accuracy on ImageNet and mCE (lower is better) on ImageNet-C between uniformly and non-uniformly allocating TAP. Taking FAN-B-Hybrid as the baseline, allocating TAP into every block yields better results than non-uniform allocation strategies in terms of both accuracy and robustness.}
  \label{tab:nonuniform_tap}%
  \vspace{-5 pt}
\end{table}%

% \subsection{Comparison of Adversarial 
% Robustness}
% \yong{
% We also evaluate the robustness against adversarial attacks. We follow the settings of RVT~\cite{mao2021towards} to construct the adversarial examples with the number of steps $t=5$ and step size $\alpha=0.5$, namely PGD-5. 
% As shown in Table~\ref{tab:adversarial}, compared to the improvement against image corruptions, the proposed methods also obtain comparable improvement against adversarial attacks. 
% }

% \begin{table}[t]
%   \begin{center}
%   \resizebox{1\columnwidth}{!}
%   {
%     \begin{tabular}{l|ccc}
%     \toprule
%     Method & \multicolumn{1}{l}{ImageNet $\uparrow$} & \multicolumn{1}{l}{ImageNet-C (mCE) $\downarrow$} & \multicolumn{1}{l}{PGD-5 $\uparrow$} \\
%     \hline
%     FAN-B-Hybrid~\cite{zhou2022understanding}   &    83.9   &  46.1     &  30.5 \\
%     ~~~+TAP   &    84.3   &  44.9 (-1.2)    &  31.4 (+0.9) \\
%     ~~~+ADL   &   84.0    &   44.4 (-1.7)    &  31.8 (+1.3) \\
%     ~~~+TAP \& ADL &    \textbf{84.3}   &   \textbf{43.7 (-2.4)}     &  \textbf{32.2 (+1.7)} \\
%     \bottomrule
%     \end{tabular}%
%   }
%   \end{center}
%   \caption{Comparisons of adversarial robustness against PGD attacks on ImageNet. We demonstrate that both our TAP and ADL also obtain promising improvement in terms of adversarial robustness. 
%   }
%   \label{tab:adversarial}%
% \end{table}%

\subsection{Effect of Attention Threshold $\tau$} \label{subsec:threshold}

\yong{
According to Eqn.~(\ref{eq:attention_threshold}), we use a threshold $\nicefrac{\tau}{N}$ {with $N$ being the number of tokens} to filter out very small values and only focus on the most important ones in the attention map via $\hat A_i^{(l)} = \mathbbm{1} (A_i^{(l)} \geq \nicefrac{\tau}{N} ) \cdot A_i^{(l)}$. Here, we explicitly study the effect of the attention threshold $\tau$ for computing our ADL.
As detailed in Table~\ref{tab:tau}, when using ADL to train the model (without TAP), our ADL only brings marginal improvements {in terms of clean performance on ImageNet. However}, our ADL becomes particularly effective in improving robustness, e.g., greatly reducing mCE on ImageNet-C. In practice, a too small or too large $\tau$ reduces the benefit of our ADL. In our experiments, we set $\tau=2$ performs to obtain the best results.
}

\begin{table}[t]
\begin{center}
  \resizebox{1\columnwidth}{!}
  {
    \begin{tabular}{c|c|cccc}
    \toprule
    $\tau$   & 0 (Baseline)  & \multicolumn{1}{c}{1} & \multicolumn{1}{c}{2} & \multicolumn{1}{c}{3} & \multicolumn{1}{c}{5} \\
    \hline
    ImageNet $\uparrow$ &   83.9    &  83.9     &    \textbf{84.0}   &   \textbf{84.0}    & 83.9 \\
    ImageNet-C (mCE) $\downarrow$ &   46.1    &   45.1 (-1.0)    &   \textbf{44.4 (-1.7)}    &  45.5 (-0.6)   &  45.8 (-0.3)  \\
    \bottomrule
    \end{tabular}%
  }
\end{center}
    \caption{Comparisons of accuracy on ImageNet and mCE (lower is better) on ImageNet-C across diverse $\tau$. We take FAN-B-Hybrid as the baseline and observe that a too small or too large $\tau$ reduces the benefit of ADL. In practice, $\tau=2$ performs best in most cases.}
  \label{tab:tau}%
\vspace{-5 pt}
\end{table}%

% \subsection{Potential Extension on Self-supervised Training}

% We agree that TAP and ADL introduce some ``manual bias''. Nevertheless, they are also general to self-supervised pretraining. \emph{\textbf{First}}, similar to self-attention, TAP can also learn by itself to determine whether to be activated or not (see discussions above). Thus, TAP is equally flexible with self-attention in self-supervised pretraining. \emph{\textbf{Second}}, ADL does not rely on any supervision signal and diversifies attention in an unsupervised manner, making it possible to be used in self-supervised training. We will explore this in the future.

\section{Conclusion}

In this paper, we address the token overfocusing issue in vision transformers (ViTs) such that ViTs tend to rely on very few important tokens in the attention mechanism. In fact, the attention is not robust and often obtains highly diverging attention patterns in the presence of corruptions. 
To alleviate this, we propose two general techniques. First, our Token-aware Average Pooling (TAP) module encourages the local neighborhood of tokens to take part in the self-attention by learning an adaptive average pooling scheme for each token. Second, our Attention Diversification Loss (ADL) explicitly reduces the cosine similarity of attention among tokens.
In practice, we apply our methods to diverse architectures and obtain a significant improvement of robustness on different benchmarks and learning tasks.

{\small
\bibliographystyle{ieee_fullname}
\bibliography{merged,survey}
}

\end{document}